\documentclass[lettersize,journal]{IEEEtran}

\usepackage[english]{babel}

\usepackage{tabularx}
\usepackage{authblk}
\usepackage{multirow}
\usepackage{ifpdf}
\usepackage[utf8]{inputenc}
\usepackage{siunitx}
\usepackage{amsmath,amsfonts}
\usepackage{algorithmic}
\usepackage{algorithm}
\usepackage{array}
\usepackage[caption=false,font=normalsize,labelfont=sf,textfont=sf]{subfig}
\usepackage{textcomp}
\usepackage{makecell}
\usepackage{stfloats}
\usepackage{url}
\usepackage{verbatim}
\usepackage{graphicx}
\usepackage{cite}
\usepackage{breqn}
\usepackage[switch]{lineno}
\usepackage{xcolor} 
\usepackage{soul}
\usepackage[colorlinks=true, allcolors=blue]{hyperref}

\begin{document}

\title{Hierarchical Information-sharing\\ Convolutional Neural Network for the Prediction of\\Arctic Sea Ice Concentration and Velocity}
\author{Younghyun Koo,~\IEEEmembership{Member,~IEEE,},
Maryam Rahnemoonfar,~\IEEEmembership{Member,~IEEE,}
\thanks{Y. Koo and M. Rahnemoonfar are with the Department of Computer Science \& Engineering and Civil \& Environmental Engineering, Lehigh University, Bethlehem, PA, 18015, USA}
}

\markboth{IEEE TRANSACTIONS ON GEOSCIENCE AND REMOTE SENSING,~Vol.~, No.~}%
{Koo \MakeLowercase{\textit{et al.}}: Multi-task CNN for Sea ice prediction}


\maketitle

\begin{abstract}
Forecasting sea ice concentration (SIC) and sea ice velocity (SIV) in the Arctic Ocean is of great significance as the Arctic environment has been changed by the recent warming climate. Given that physical sea ice models require high computational costs with complex parameterization, deep learning techniques can effectively replace the physical model and improve the performance of sea ice prediction. This study proposes a novel multi-task fully conventional network architecture named hierarchical information-sharing U-net (HIS-Unet) to predict daily SIC and SIV. Instead of learning SIC and SIV separately at each branch, we allow the SIC and SIV layers to share their information and assist each other's prediction through the weighting attention modules (WAMs). Consequently, our HIS-Unet outperforms other statistical approaches, sea ice physical models, and neural networks without such information-sharing units. The improvement of HIS-Unet is more significant to when and where SIC changes seasonally, which implies that the information sharing between SIC and SIV through WAMs helps learn the dynamic changes of SIC and SIV. The weight values of the WAMs imply that SIC information plays a more critical role in SIV prediction, compared to that of SIV information in SIC prediction, and information sharing is more active in marginal ice zones (e.g., East Greenland and Hudson/Baffin Bays) than in the central Arctic.
\end{abstract}

\begin{IEEEkeywords}
Machine learning, Sea ice forecast, Information sharing, weighting attention module, Sea ice motion, Cryosphere, U-net 
\end{IEEEkeywords}

\section{Introduction}
\IEEEPARstart{T}{he} Arctic sea ice extent and thickness have changed dramatically over the last few decades. Sea ice extent (SIE) has been reduced by more than \SI{50000}{\km\squared}/year (4 \%/decade) since the era of satellite observation in the 1970s \cite{Cavalieri2012, Stroeve2012}, which is mainly attributed to anthropogenic CO2 emission and resultant global warming \cite{Notz2016}. Such a loss of the Arctic sea ice cover has been observed for all seasons, both winter and summer, and almost all regions in the Arctic Ocean \cite{Onarheim2018}. Meanwhile, the Arctic sea ice thickness (SIT) has decreased by more than 2 m ($>$ 60 \%), and more than half of multi-year ice (MYI) has disappeared for the last few decades \cite{Lindsay2015, Kwok2018}. Such dramatic changes in SIE and SIT could have affected the thermodynamic and dynamic conditions of the Arctic sea ice \cite{Yu2020, Kwok2021, Auheuser2023}. In particular, since the dynamic movement of sea ice has significant impacts on sea ice area and sea ice volume, it is important to understand both sea ice dynamics and area fraction together as a clue in the future Arctic and global climate \cite{Wagner2021}.

To predict the Arctic sea ice motions, a number of numerical sea ice models have been developed based on a physical understanding of sea ice and its interaction with the atmosphere and ocean. However, such physics models require too many complicated parameterizations \cite{Blockley2020, hunke2010} accompanying high computational costs to run the models. Moreover, considering that numerical models are highly sensitive to initial conditions and physical assumptions, they can produce prediction results inconsistent with real observations \cite{Wrigglesworth2015}.

Recently, in addition to these physical numerical models, machine learning (ML) techniques, particularly deep learning, have emerged as another efficient way to forecast sea ice conditions. The most common sea ice variable predicted by the deep learning model is sea ice concentration (SIC). Myriad deep learning models, including convolutional neural network (CNN) and recurrent neural network (RNN), have been proposed to predict SIC \cite{Andersson2021, Kim2020, Ren2021, Ren2022, Wei2022} with lead times varying from daily \cite{Liu2021_daily, Zheng2022, Liu2021_2} to weekly \cite{Liu2021_weekly} and monthly scale \cite{Chi2017, Mu2023}. While most previous deep learning studies have focused on predicting SIC, deep learning has rarely been used for predicting sea ice velocity (SIV). Nevertheless, several studies demonstrated that CNN outperformed other statistical methods in daily SIV prediction \cite{Hoffman2023, zhai2021}.

However, those previous ML studies have only focused on the prediction of a single variable, such as either SIC or SIV alone. Considering that SIC and SIV affect each other, these two variables can potentially complement each other to improve their own prediction. Thus, this study develops a fully convolutional deep learning model performing short-term forecasting of SIC and SIV simultaneously. We design a fully convolutional network (FCN) architecture consisting of two branches specialized for SIC and SIV prediction, which share their intermediate prediction layers through information-sharing blocks. We aim to improve both SIC and SIV prediction performances by allowing the SIC and SIV intermediate layers to share their information during the model training.

The main contributions of this research work consist of the following.
\begin{itemize}
    \item We design a novel FCN architecture called Hierarchical information-sharing U-net (HIS-Unet) to predict SIC and SIV simultaneously. 
    \item To allow sharing and emphasizing important information between SIC and SIV, we insert weighting attention modules (WAMs) between SIC and SIV branches, consisting of linear weights to SIC and SIV features and channel \& spatial attention modules.
    \item We conduct extensive experiments to check if sharing and highlighting information between SIC and SIV branches via WAMs improves SIC and SIV predictability and when and where these improvements are significant.
    \item By interpreting the spatial patterns of information sharing at WAMs, we exhibit how the SIC and SIV information is shared for different regions.
    \item By comparing the climatological conditions of the test year with training years, we check whether sharing SIC and SIV information can assist each other's prediction even in climatologically anomalous years.
\end{itemize}

The remainder of the paper is organized as follows. Section \ref{related_work} reviews several literature regarding the physical models for sea ice prediction, machine learning for sea ice prediction, and multi-task machine learning. Section \ref{data} explains details of remote sensing and meteorological data used in this study, and section \ref{method} presents the detailed architecture of our information-sharing network and the baseline models. The accuracy and implication of the model are given in Section \ref{results} and \ref{discussion}, respectively.

\section{Related work}\label{related_work}

In the following subsections, we first discuss the prediction of sea ice dynamics with physical models. Next, we discuss the recent development of neural network techniques for sea ice prediction. Finally, we discuss the recent development of multi-task neural networks in remote sensing.

\subsection{Dynamic sea ice models}\label{sea_ice_dynamics}
In general, changes in SIC are driven by two components: dynamic and thermodynamic processes. The evolution of SIC ($A$) is governed by the following equation \cite{Holland2012}:
\begin{equation}\label{eq:1}
\frac{\partial A}{\partial t} + \nabla\cdot(\boldsymbol{u}A)=f_c-r 
\end{equation}
where $\boldsymbol{u}$ is ice motion, $f_c$ is SIC change from freezing or melting (thermodynamic process), and $r$ is SIC change from mass-conserving mechanical ice redistribution processes (e.g., ridging or rafting) that convert ice area to ice thickness. Based on this relationship, the SIC changes in the Arctic and Southern Oceans have been explored as the combination of sea ice motion, thermodynamic growth, and mechanical ice deformation \cite{Holland2012, Holland2016}.

Additionally, numerous physical sea ice models have been proposed to explain and predict the dynamic behavior of the Arctic sea ice. Assuming the elastic-viscous-plastic (EVP) condition of sea ice, these models are governed by the following momentum equation \cite{Hibler1979}:
\begin{equation}\label{eq:3}
m\frac{D \boldsymbol{u}}{D t} = -mf\boldsymbol{k}\times\boldsymbol{u} + \tau_{ai} + \tau_{wi} + \boldsymbol{F} -mg \nabla H
\end{equation}
where $D/Dt=\partial/\partial t + \boldsymbol{u} \cdot \nabla$ is the substantial time derivative, $m$ is the ice mass per unite area, $\boldsymbol{k}$ is a unit vector normal to the surface, $\boldsymbol{u}$ is the ice velocity, $f$ is the Coriolis parameter, $\tau_{ai}$ and $\tau_{wi}$ are the forces due to air and water stresses, $H$ is the elevation of the sea surface, $g$ is the gravity acceleration, and $\boldsymbol{F}$ is the force due to variations in internal ice stress. As many previous studies have already suggested, wind and ocean forcings have primary impacts on SIV. Particularly, wind velocity has been treated as a major variable in SIV, which can contribute to up to 70 \% of the sea ice velocity variances \cite{Thorndlike1982} depending on season or region \cite{Maeda2020}. Nevertheless, predicting sea ice dynamics based on physical models is still challenging due to its intrinsic complexity and dependency on numerous atmospheric and oceanic parameterizations.

\subsection{Deep learning for sea ice prediction}\label{NN_for_seaice}

Nowadays, various ML and deep learning techniques have been extensively used in many applications of remote sensing of polar oceans \cite{Baek2023, huang2024remote, Stokholm2022, li2020deep, Gao2023}. In particular, deep learning techniques have been effective in predicting the high complexity of sea ice conditions (i.e., SIC and SIV) \cite{Li2024_RS}. First, in terms of SIC, Andersson et al. \cite{Andersson2021} proposed a deep-learning sea ice forecasting system named IceNet, which is designed to forecast monthly SIC for the next six months. Kim et al. \cite{Kim2020} also used CNN to predict after-one-month SIC from satellite-based SIC observations and weather data for the previous months, and their model showed a mean absolute error (MAE) of 2.28\% and anomaly correlation coefficient of 0.98. Similarly, a CNN model proposed by Ren and Li \cite{Ren2021} showed $<$ 1 \% of MAE in daily SIC prediction for melting season, and U-Net by Ren et al. \cite{Ren2022} showed 2-3 \% of MAE in daily SIC prediction. In addition to CNN, long short-term memory (LSTM), an advanced RNN architecture, has been commonly used to predict SIC. LSTM models generally have shown $>$ 0.98 correlation coefficient and $<$ 10 \% of root mean square error (RMSE) for a daily-scale SIC prediction \cite{Liu2021_daily, Liu2021_2}. The monthly SIC predictions performed by LSTM models have shown $<$ 10-12 \% of RMSE \cite{Chi2017, Mu2023, Wei2022}.

CNN and LSTM have also been used for SIV prediction in several studies. Zhai et al. \cite{zhai2021} used CNN to predict SIV, and their model showed a correlation coefficient $>$ 0.8, outperforming other statistical and physical models. Similarly, the CNN model proposed by Hoffman et al. \cite{Hoffman2023} showed approximately 0.8 correlation coefficient in SIV prediction. The convolutional LSTM presented by Petrou and Tian \cite{Petrou2019} showed 2.5-3.1 km/day of MAE and 3.7-4.2 km/day of RMSE in SIV prediction. However, to our knowledge, most previous machine learning studies only focused on deriving either SIC or SIV, not attempting to integrate SIC and SIV information. This study is the first to integrate SIC and SIV information in a single multi-task neural network architecture, aiming to improve the prediction of both variables.

\begin{table}
\centering
\caption{Examples of deep learning for sea ice prediction}
\begin{tabular}{ccc}
\hline
Sea ice variable & Method & References\\
\hline
\multirow{3}{*}{Sea ice concentration} & CNN & \cite{Andersson2021, Kim2020, Ren2021, Ren2022} \\
 & LSTM & \cite{Zheng2022, Wei2022, Chi2017, Mu2023}\\
  & Conv-LSTM & \cite{Liu2021_daily, Liu2021_weekly, Liu2021_2}\\
\hline
\multirow{2}{*}{Sea ice velocity} & CNN & \cite{zhai2021, Hoffman2023}\\
 & LSTM & \cite{Petrou2019}\\
\hline

\end{tabular}
\label{table_ML}
\end{table}

\subsection{Multi-task neural network}\label{MTNN}
In the existing multi-task network studies, the most common architecture is the ``shared-trunk'': all tasks share certain hidden layers to learn common features for all tasks but keep several task-specific branched layers \cite{Crawshaw2020}. Conventionally, multi-task networks can be categorized into two formulations depending on how to choose the branched level and share the information between tasks: (1) early-branched network and (2) late-branched network. The early-branched or late-branched approaches have been adopted in various fields for the classification and segmentation of remote sensing imagery data \cite{Alhichri2018, Papadomanolaki2021, Ilteralp2022, Li2023_RS}. Alhichri \cite{Alhichri2018} used a multi-task architecture similar to a late-branch network for multiple land-use classification tasks. Papadomanolaki et al. \cite{Papadomanolaki2021} proposed a UNet-like multi-task architecture to perform segmentation and change detection from multi-spectral image sets. Ilteralp et al. \cite{Ilteralp2022} used a multi-task convolutional network for Chlorophyll-a estimation and month classification with satellite images. Li et al. \cite{Li2023_RS} proposed a multi-task network to perform mask prediction, edge prediction, and distance map estimation. 


However, this shared-trunk architecture makes a model inflexible, so the model performance highly depends on where different branches split up. A late-branch network can lose significant task-specific features because nearly all network parameters are shared across all training processes \cite{He2021}. Excessive sharing can also hinder the performance of another task that has different needs (a.k.a. negative transfer). On the contrary, an early-branched network can fail to leverage information effectively between tasks because of too little sharing \cite{Crawshaw2020}. Hence, instead of the shared trunk architecture, another \emph{cross-talk} approach has been proposed: each task has a separate network, but information flow is added between parallel layers of the task networks. For example, Misra et al. \cite{Misra2016} proposed cross-stitch units between separate CNNs for different tasks, which provide the optimal linear combinations for a given set of tasks. Similarly, a sluice meta-network introduced by Ruder et al. \cite{Ruder2019} consists of a shared input layer, two task-specific output layers, and three hidden layers that have task-specific and shared subspaces. The input of each layer is determined as a linear combination of task-specific and shared outputs of the previous layers. Instead of a linear combination of the additional parameters, Gao et al. \cite{Gao2019} used $1\times1$ convolution to perform neural discriminative dimensionality reduction (NDDR) to retain the discriminative information from task-specific and shared features. Similarly, He et al. \cite{He2021} introduced task consistency learning (TCL) blocks in their hierarchically fused U-Net. However, to the best of our knowledge, no studies have conducted such a multi-task network for sea ice applications. This paper proposes the first multi-task network designed for SIC and SIV prediction.


\section{Data}\label{data}

We collect SIV and SIC satellite observation data as the input and output of the models. We also collect wind and air temperature data from the ERA5 climate reanalysis product as additional input variables. Sea ice physics model data is also collected to be used for a baseline comparison. We collect the data from 2016 to 2022 and filter out any data with null values or physically invalid values. The following subsections provide details of the datasets we use (Table \ref{table_data}).

\subsection{Sea ice velocity}
For the SIV data, we use the NSIDC Polar Pathfinder Daily 25 km EASE-Grid Sea Ice Motion Vectors version 4 \cite{NSIDC, Tschudi2020}. This product derives daily sea ice drift from three primary types of sources: (1) gridded satellite imagery (e.g., AVHRR, SSMI, SSMI/S, AMSR), (2) NCEP/NCAR wind reanalysis data, and (3) buoy position data from the International Arctic Buoy Program (IABP). The u component (along-x) and v component (along-y) of sea ice motions are independently derived from each of these sources and optimally interpolated onto a 25 km Equal-Area Scalable Earth (EASE) grid by combining all sources. When sea ice drift is derived from satellite image data, a correlation coefficient is calculated between a small target area in one image and a searching area in the second image. Then, the location in the second image where the correlation coefficient is the highest is determined as the displacement of ice \cite{NSIDC}. The mean difference between the interpolated u components and the buoy vectors was approximately 0.1 km/day with a Root Mean Square (RMS) error of 2.9 km/day, and 0.3 km/day with an RMS error of 2.9 km/day for v components \cite{NSIDC}. The SIV extraction schemes of this dataset are valid at significant distances from the ice edges, in areas where ice conditions are relatively stable, stationary, homogenous, and isotropic day to day. Therefore, surface melting in the summer season can deteriorate drift accuracy because it affects the passive-microwave identification of ice parcels \cite{Tschudi2020}. In this study, we use SIV values at least 50 km (or 2 pixels) away from the coastline.


\subsection{Sea ice concentration}
For the SIC data, we use NOAA/NSIDC Climate Data Record of Passive Microwave Sea Ice Concentration version 4 data \cite{Meier2021}. This data set provides a Climate Data Record (CDR) of SIC (i.e., the areal fraction of ice within a grid cell) from passive microwave (PMW) data, including SMMR, SSMI, SSMI/S. The CDR algorithm output is the combination of SIC estimations from two algorithms: the NASA Team (NT) algorithm \cite{NASA_team} and NASA Bootstrap (BT) algorithm \cite{NASA_BT}. These empirical algorithms estimate SIC from the PMW brightness temperatures at different frequencies and polarizations (i.e., vertical and horizontal polarizations at 19 GHz, 22 GHz, and 37 GHz). Then, the CDR product adjusts algorithm coefficients for each sensor to optimize the consistency of daily and monthly SIC time series. Several assessments showed that the error of this SIC estimation is approximately 5 \% within the consolidated ice pack during cold winter conditions \cite{Meier2005, Comiso1997, Ivanova2015}. However, in the summer season, the error can rise to more than 20 \%  due to surface melt and the presence of melt ponds \cite{Kern2020}. Due to the data quality issue near coastal areas, we use the SIC data more than 50 km from the coastline. Since the projection of this SIC data (NSIDC Sea Ice Polar Stereographic North) is different from that of SIV data (EASE grid), we reproject SIC into the EASE grid using bilinear interpolation.

\begin{table*}
\centering
\caption{Input data sets for machine learning models}
\begin{tabular}{cccc}
\hline
Dataset & Name& \makecell{Spatial\\resolution} & \makecell{Temporal\\resolution}\\
\hline
Sea ice velocity (u and v)& NSIDC Polar Pathfinder Daily 25 km EASE-Grid Sea Ice Motion Vectors \cite{Tschudi2020} & 25 km & Daily\\
Sea Ice Concentration & NOAA/NSIDC Climate Data Record of Passive Microwave Sea Ice Concentration \cite{Meier2021} & 25 km & Daily\\
Wind velocity (u and v) & ECMWF Reanalysis v5 (ERA5) hourly data on single levels \cite{ERA5} & 0.25$^{\circ}$ & Hourly\\
Air temperature & ECMWF Reanalysis v5 (ERA5) hourly data on single levels \cite{ERA5} & 0.25$^{\circ}$ & Hourly\\
X Y coordinates & 25 km Equal-Area Scalable Earth (EASE) Grid & 25 km & -\\
\hline
\end{tabular}
\label{table_data}
\end{table*}%

\subsection{ERA5 climate reanalysis}
ERA5 is the fifth generation ECMWF (European Centre for Medium-Range Weather Forecasts) atmospheric reanalysis of the global climate covering the period from January 1940 to the present \cite{ERA5}. ERA5 provides hourly estimates of atmospheric, land, and oceanic climate variables \cite{ERA5}. We acquire the daily average wind velocity (u and v components) at 10 m height and 2 m air temperature from this hourly data. In addition, since the raw ERA5 data is gridded to a regular latitude-longitude grid of 0.25 degrees, we reproject this data onto the 25 km EASE grid using bilinear interpolation to co-locate with the SIV data.

\subsection{Sea ice physics model}
As a baseline to compare the performance of our ML model, we use the Arctic Ocean physics analysis and forecast data (Level 4) provided by Copernicus Marine Services. This data uses the operational TOPAZ4 Arctic Ocean system \cite{TOPAZ4} with Hybrid Coordinate Ocean Model (HyCOM) \cite{HYCOM}. It is run daily to provide 10 days of forecast (average of 10 members) of the 3D physical ocean, including sea ice albedo, sea ice area fraction, sea ice thickness, and sea ice velocity. The original gridded data (12.5 km resolution at the North Pole on a polar stereographic projection) is also converted to the 25 km EASE grid.

\section{Method}\label{method}

Given that the interaction between SIC and SIV is extremely complicated by various thermodynamic and dynamic mechanisms, the multi-task model for SIC and SIV prediction should be flexible and interactive. Hence, in order to share SIC and SIV information efficiently, we adopt the cross-talk multi-task architecture \cite{Crawshaw2020} instead of the shared-trunk approach. Additionally, based on the assumption that sharing SIC and SIV information throughout the learning process improves their prediction, we add weighting attention modules (WAM) between the SIC and SIV task-specific networks to allow sharing and highlighting of essential information. The details of the proposed multi-task network architecture and comparison with baseline models are described in the following subsections.

\begin{figure*}
    \centering
    \includegraphics[width=0.92\textwidth]{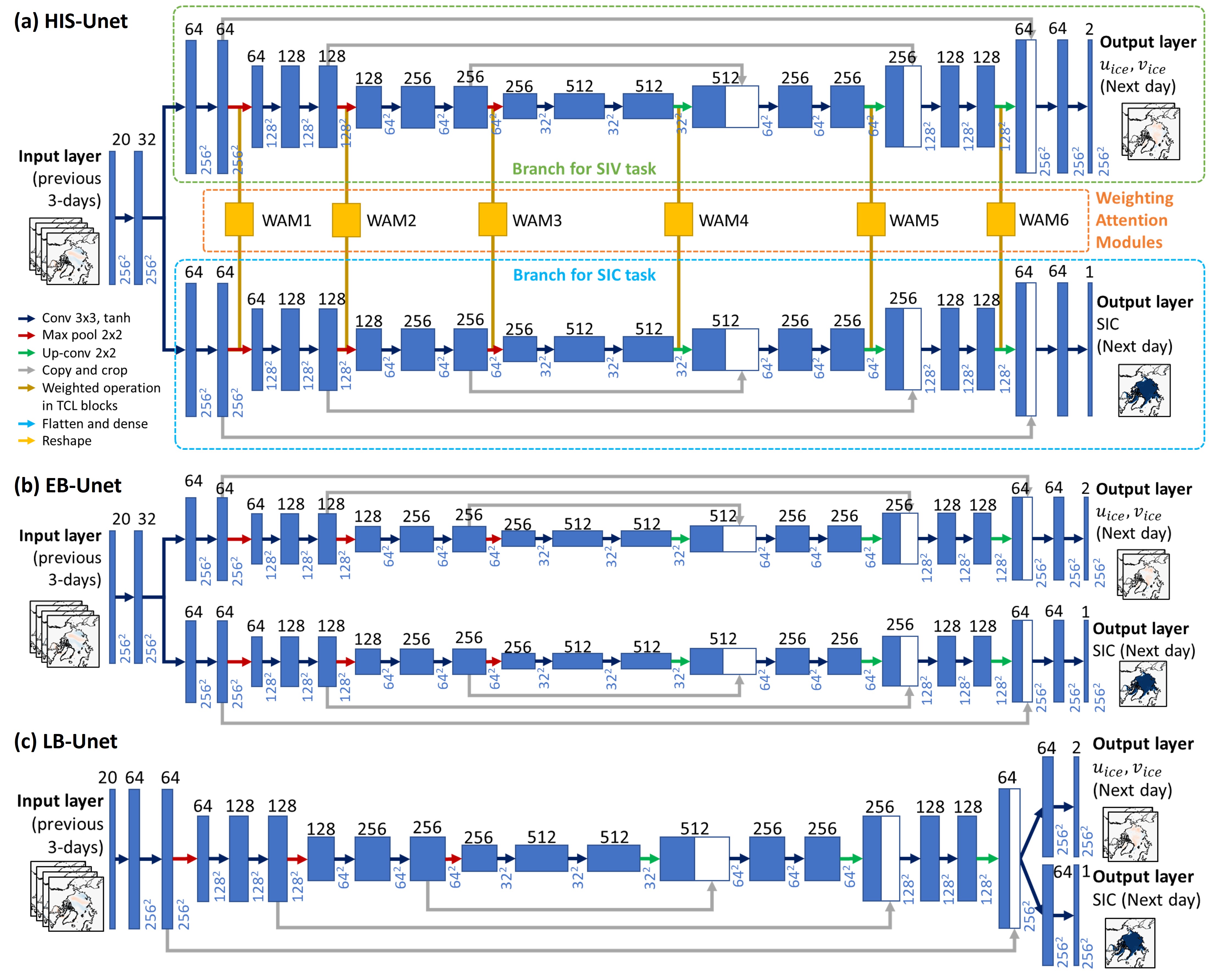}
    \caption{Architecture of multi-task CNN models tested in this study: (a) Hierarchical information-sharing U-net (HIS-Unet), (b) early-branched U-net (EB-Unet), and (c) late-branched U-net (LB-Unet).}
    \label{Model_architectures}
\end{figure*}

\begin{figure}
    \centering
    \includegraphics[width=0.9\linewidth]{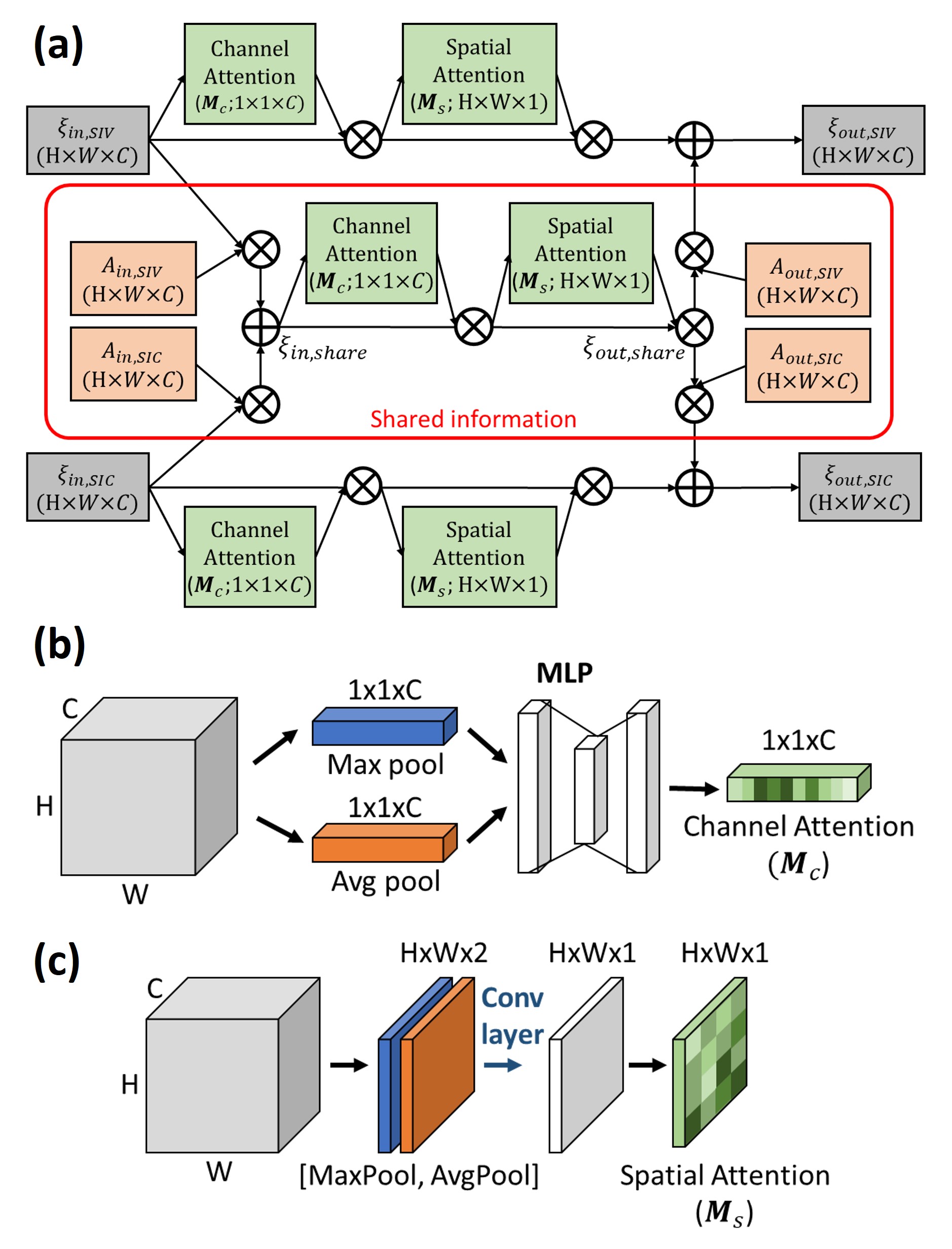} 
    \caption{Schematic diagram of the (a) Weighting attention module (WAM), (b) Channel attention module, and (c) Spatial attention module.}
    \label{WAM_block}
\end{figure}

\subsection{Hierarchical information-sharing U-net}

Fig. \ref{Model_architectures}a describes the architecture of the multi-task network we propose in this study: Hierarchical information-sharing U-net (HIS-Unet). In this architecture, two separate SIV and SIC prediction tasks have their own branches after a common convolutional layer of 32 filters. Each branch has a U-net structure \cite{Ronneberger2015} consisting of the encoder (contracting path) and decoder (expansive path). Each encoder consists of repeated two $3\times3$ convolutions and a $2\times2$ max pooling operation with stride 2 for downsampling. The number of feature channels is doubled after each downsampling step. The decoder consists of several upsampling steps implemented by a $2\times2$ up-convolution that halves the number of feature channels, a concatenation with the cropped feature map, and two $3\times3$ convolutions \cite{Ronneberger2015}. The hyperbolic tangent (tanh) activation function is applied after each convolutional layer. Although each branch has a separate U-net construction that produces SIV or SIC, they share and transfer their information through six weighting attention modules (WAMs). By adopting linear weighting parameters and channel/spatial attention modules in WAMs, we enable these WAMs to (1) determine how much information from SIC and SIV is shared with each other and (2) highlight important channel and spatial information. We insert six WAMs right after the max pooling or up-convolution operations: three blocks in the encoder steps and three blocks in the decoder steps.

As shown in Fig. \ref{WAM_block}a, each WAM first receives information from SIV and SIC branches and calculates the weighted sum of them. Letting a WAM receive the SIV feature map ($\xi_{in,SIV}$; image height $H$, image width $W$, channels $C$) and SIC feature map ($\xi_{in,SIC}$; $H \times W \times C$) from the previous max-pooling or up-convolution operations, the input shared information ($\xi_{in,share}$) is determined by following:
\begin{equation}\label{eq:4}
\xi_{in,share} = A_{in,SIV}\xi_{in,SIV} + A_{in,SIC}\xi_{in,SIC}
\end{equation}
where $A_{in,SIV}$ and $A_{in,SIC}$ denote the weights for the SIV and SIC information, respectively. We initially set both $A_{in,SIV}$ and $A_{in,SIC}$ as 0.5, which indicates both SIV and SIC information are equally shared before training the model. As the model is trained with data, $A_{in,SIV}$ and $A_{in,SIC}$ are fine-tuned for the best accuracy. If $A_{in,SIV}$ is high, more SIV information is shared during the information sharing; if $A_{in,SIC}$ is high, more SIC information is shared.

After the input shared information $\xi_{in,share}$ is determined by the linear weighting combination of $\xi_{in,SIV}$ and $\xi_{in,SIC}$, this shared information passes through channel attention module (Fig. \ref{WAM_block}b) and spatial attention module (Fig. \ref{WAM_block}c). The channel and spatial attention module were proposed by Woo et al. \cite{Woo2018} and have been widely used to improve the representation power of CNNs. The channel attention highlights what channel is meaningful in a given input image, and the spatial attention highlights where an informative part is located \cite{Woo2018}. In the channel attention module, the spatial information of an input feature map is aggregated using max pooling and average pooling operations, and these two descriptors are forwarded to a multi-layer perceptron (MLP) with one hidden layer. Consequently, the channel attention map of $1 \times 1 \times C$ dimension is generated, and this channel attention map is multiplied with the input feature of $H \times W \times C$ dimension. After the channel attention is applied, the spatial attention is applied. In the spatial attention module, average pooling and max pooling are applied along the channel axis, and these two feature descriptors generate a spatial attention map ($H \times W \times C$ dimension) through a convolutional layer. As a result, these channel and spatial attention modules convert the input shared information $\xi_{in,share}$ into the output shared information ($\xi_{out,share}$):
\begin{equation}\label{eq:5}
\xi_{out,share} = \boldsymbol{M}_s(\xi_{in,share}) \otimes (\boldsymbol{M}_c(\xi_{in,share}) \otimes \xi_{in,share})
\end{equation}
where $\boldsymbol{M}_s$ and $\boldsymbol{M}_c$ denote spatial attention map and channel attention map derived from spatial and channel attention modules, respectively, and $\otimes$ denotes element-wise multiplication. If a certain channel is important in model training, this channel should have a higher channel attention value than the others. Similarly, if a specific spatial coordinate is important, this location is highlighted as higher value in the spatial attention. By arranging the channel attention and spatial attention sequentially, we can highlight what and where shared information is important. These channel and spatial attention modules are also applied to the input SIV and SIC feature maps ($\xi_{in,SIV}$ and $\xi_{in,SIC}$).

Then, the attention shared information is sent to the SIV and SIC branches after multiplying output weights ($A_{out,SIV}$ and $A_{out,SIC}$) and adding attention SIV and SIC information, respectively:
\begin{equation}\label{eq:6_1}
\begin{aligned}
\xi_{out,SIV} &= A_{out,SIV}\xi_{out,share} \\
&+ \boldsymbol{M}_s(\xi_{in,SIV}) \otimes (\boldsymbol{M}_c(\xi_{in,SIV}) \otimes \xi_{in,SIV})
\end{aligned}
\end{equation}
\begin{equation}\label{eq:6_2}
\begin{aligned}
\xi_{out,SIC} &= A_{out,SIC}\xi_{out,share} \\
&+ \boldsymbol{M}_s(\xi_{in,SIC}) \otimes (\boldsymbol{M}_c(\xi_{in,SIC}) \otimes \xi_{in,SIC})
\end{aligned}
\end{equation}
Here, $A_{out,SIV}$ and $A_{out,SIC}$ determine the relative importance of shared information to each branch. A higher $A_{out,SIV}$ value means that more shared information is added to the SIV branch, implying that SIC information is helpful for SIV prediction. Likewise, a higher $A_{out,SIC}$ value corresponds to a greater contribution of SIV information to the SIC prediction. In section \ref{discussion2}, we will discuss details about how those weights in the WAMs affect the SIC and SIV predictions.


The objective loss function we use for learning is the sum of the mean square error (MSE) for u-component SIV ($u$), v-component SIV ($v$), and SIC ($A$):
\begin{equation}\label{eq:7}
Loss = \sum (\lvert u_{prd}-u_{obs} \rvert^2 + \lvert v_{prd}-v_{obs} \rvert^2 + \beta\lvert A_{prd}-A_{obs} \rvert^2)
\end{equation}
where the subscript $obs$ means observation and $prd$ means prediction, and $\beta$ is the weight for SIC MSE. When we experimented with $\beta$ values of 0.1, 0.2, 0.5, and 1.0, $\beta = 0.5$ produced the best accuracy, balancing both SIV and SIC predictability. Thus, we set $\beta$ to 0.5.

\subsection{Baseline models}

To compare with the HIS-Unet, we evaluate five neural network models and two simple statistical models. Five neural network models are (1) Early-branched U-net (EB-Unet), (2) Late-branched U-net (LB-Unet), (3) U-net, (4) FCN only with convolutional layers, and (5) CNN proposed by Hoffman et al. \cite{Hoffman2023} for SIV; two statistical models are (1) Persistence and (2) Linear regression (LR) models.

Based on the U-net architecture \cite{Ronneberger2015}, we test two multi-task networks: early-branched U-net (EB-Unet) (Fig. {\ref{Model_architectures}}b) and late-branched U-net (LB-Unet) (Fig. {\ref{Model_architectures}}c). The EB-Unet only shares the first convolutional layer, but the rest of the convolutional layers are separated into SIV and SIC branches of independent U-net architecture. On the other hand, the LB-Unet has the same architecture as the original U-net, but the last convolutional layer and output layer are branched for SIV and SIC prediction separately. In addition to these multi-task U-nets, we also test the original U-net architecture \cite{Ronneberger2015} with three channels of u-component SIV, v-component SIV, and SIC at the output layer.

Another neural network model we evaluate is a simple FCN consisting of 7 convolutional layers of 64 filters for each and an output layer with three channels of u-component SIV, v-component SIV, and SIC. We refer to this network as FCN7. The final neural network we evaluate is the CNN architecture proposed by Hoffman et al. \cite{Hoffman2023}. This model consists of five sequential layers of convolutional and max pooling, followed by the flattening, dense, and reshaping layers. Since the original CNN model \cite{Hoffman2023} only predicts u- and v- component SIV, we add an SIC output channel at the output layer.

As a statistical model, the persistence model simply assumes that the sea ice condition remains the same as that of the previous day. The LR model, another statistical model, predicts $u$, $v$, and $A$ at pixel $(i,j)$ from the linear combination of the input variables as follows:
\begin{equation}\label{eq:8}
u(i,j)=\sum_{k=1}^{n}a_{u,k}(i,j) x_{k}(i,j)
\end{equation}
\begin{equation}\label{eq:9}
v(i,j)=\sum_{k=1}^{n}a_{v,k}(i,j) x_{k}(i,j)
\end{equation}
\begin{equation}\label{eq:10}
A(i,j)=\sum_{k=1}^{n}a_{A,k}(i,j) x_{k}(i,j)
\end{equation}
where $n$ is the number of input variables and $x_{k}$ is the $k$th input variable. $a_{u,k}$, $a_{v,k}$, and $a_{A,k}$ are the linear coefficients corresponding to $x_{k}$ for the prediction of $u$, $v$, and $A$, respectively.

\subsection{Training and testing models}
To predict the daily SIV and SIC, we use the previous 3-days of SIV (u- and v- components), SIC, air temperature, wind velocity (u- and v- components), and geographic location (i.e., x and y coordinates) as the inputs of the prediction models. Consequently, the input layer has 20 channels (three-day data of six sea ice or meteorological variables and two geographical variables) of $256\times256$ grid size. All input values are normalized to -1 to 1 based on the nominal maximum and minimum values that each variable can have. Then, this model is trained to predict the SIV and SIC for the next day. We collect the data from 2016 to 2022; 2016-2021 datasets are used to train/validate the model, and 2022 datasets are used to test the model. The 2016-2021 datasets are randomly divided into 80 \% of train datasets and 20 \% of validation datasets. The number of train, validation, and test datasets is 1742, 438, and 363, respectively. The HIS-Unet is trained to minimize the loss function in Eq.\ref{eq:7} and optimized by Adam stochastic gradient descent algorithm \cite{Kingma2015} with 100 epochs and 0.001 learning rate. Such hyperparameter settings are the same for the other neural network models. We implement this model in Python using the PyTorch library. All scripts are executed on the GPU nodes of the Frontera supercomputing system from the Texas Advanced Computing Center (TACC), each node equipping four NVIDIA Quadro RTX 5000 GPUs of 16 GB memory.

\subsection{Evaluation metrics}

The performances of the neural network models, statistical baseline models, and sea ice physics model are evaluated by the following three metrics: (1) Correlation coefficient (R), (2) RMSE, and (3) MAE:
\begin{equation}\label{eq:11}
\text{R}=\frac{\sum(x-\overline{x})(y-\overline{y})}{\sqrt{\sum(x-\overline{x})^2 \sum(y-\overline{y})^2}}
\end{equation}
\begin{equation}\label{eq:12}
\text{RMSE}=\sqrt{\frac{(x-y)^2}{N}}
\end{equation}
\begin{equation}\label{eq:13}
\text{MAE}=\sum\frac{\lvert x-y \rvert}{N}
\end{equation}
where $x$ denotes the prediction values, $y$ denotes the observation values, and $N$ is the number of samples. In calculating these metrics for SIV, we calculate the average of u-component and v-component SIV. In general, R can be useful to assess the overall spatiotemporal pattern of SIC and SIV variations, and RMSE or MAE can be useful to assess the magnitude of prediction errors.

\section{Results}\label{results}

\begin{table*}
\centering
\caption{Accuracy of SIC and SIV prediction for each model}
\begin{tabular}[t]{l|rrr|rrr}
\hline
\multirow{2}{*}{Model} &
\multicolumn{3}{c|}{Sea ice concentration (SIC)} &
\multicolumn{3}{c}{Sea ice velocity (SIV)} \\
&MAE (\%) &RMSE (\%) &R & MAE (km/d) &RMSE (km/d) &R\\
\hline
HIS-Unet &\textbf{2.934} &\textbf{6.122} &\textbf{0.978} &\textbf{1.812} &\textbf{2.667}	&\textbf{0.834} \\
EB-Unet &3.139 &6.238 &0.974 &1.926 &2.773 &0.818 \\
LB-Unet &3.122 &6.575 &0.970 &1.960 &2.814 &0.813 \\
Unet &3.400 &6.601 &0.970 &1.981 &2.834 &0.815 \\
FCN7	&4.750 &7.896 &0.958 &2.331 &3.213 &0.770 \\
CNN \cite{Hoffman2023} &10.071 &16.963 &0.839 &2.223 &3.217 &0.742\\
LR &4.232 &8.178 &0.965 &3.340 &5.142 &0.780 \\
Persistence	&3.135 &7.435 &0.924 &2.871 &4.181 &0.629 \\
HYCOM \cite{HYCOM} &8.874 &15.878 &0.865 &3.782 &5.506 &0.508 \\
\hline
\end{tabular}
\label{table2}
\end{table*}%

Table \ref{table2} shows the test accuracy of SIC and SIV in 2022 for different models. The HIS-Unet model generally shows the best SIC and SIV prediction results, with the highest R (0.978 for SIC and 0.834 for SIV) and lowest RMSE (6.122 \% for SIC and 2.677 km/day for SIV), followed by EB-Unet, LB-Unet, Unet, and FCN7. These five fully convolutional neural network models (HIS-Unet, EB-Unet, LB-Unet, Unet, and FCN7) generally show better performance than the persistence model for both SIC and SIV, suggesting that these models can significantly predict the daily variations of sea ice conditions. Additionally, these machine learning models outperform the LR model and HYCOM physical model. It is also worth mentioning that the CNN proposed by Hoffman et al. \cite{Hoffman2023} has better fidelity than the persistence model in SIV but the worst performance in SIC prediction. Since this CNN model was originally designed for only SIV prediction, this architecture does not produce robust predictions for SIC.

The EB-Unet shows considerable improvement in both SIV and SIC prediction compared to LB-Unet and U-net. This implies that separating SIC and SIV information at the early-branch stage is more efficient in predicting both variables than using the late-branch architecture or adding channels at the output layer. However, considering that such an early-branch approach prevents information sharing at the intermediate learning stages, the information sharing between SIC and SIV through WAMs in the middle of HIS-Unet architecture allows a better prediction for both SIC and SIV than the EB-Unet.

\begin{figure}
    \centering
    \includegraphics[width=1.0\linewidth]{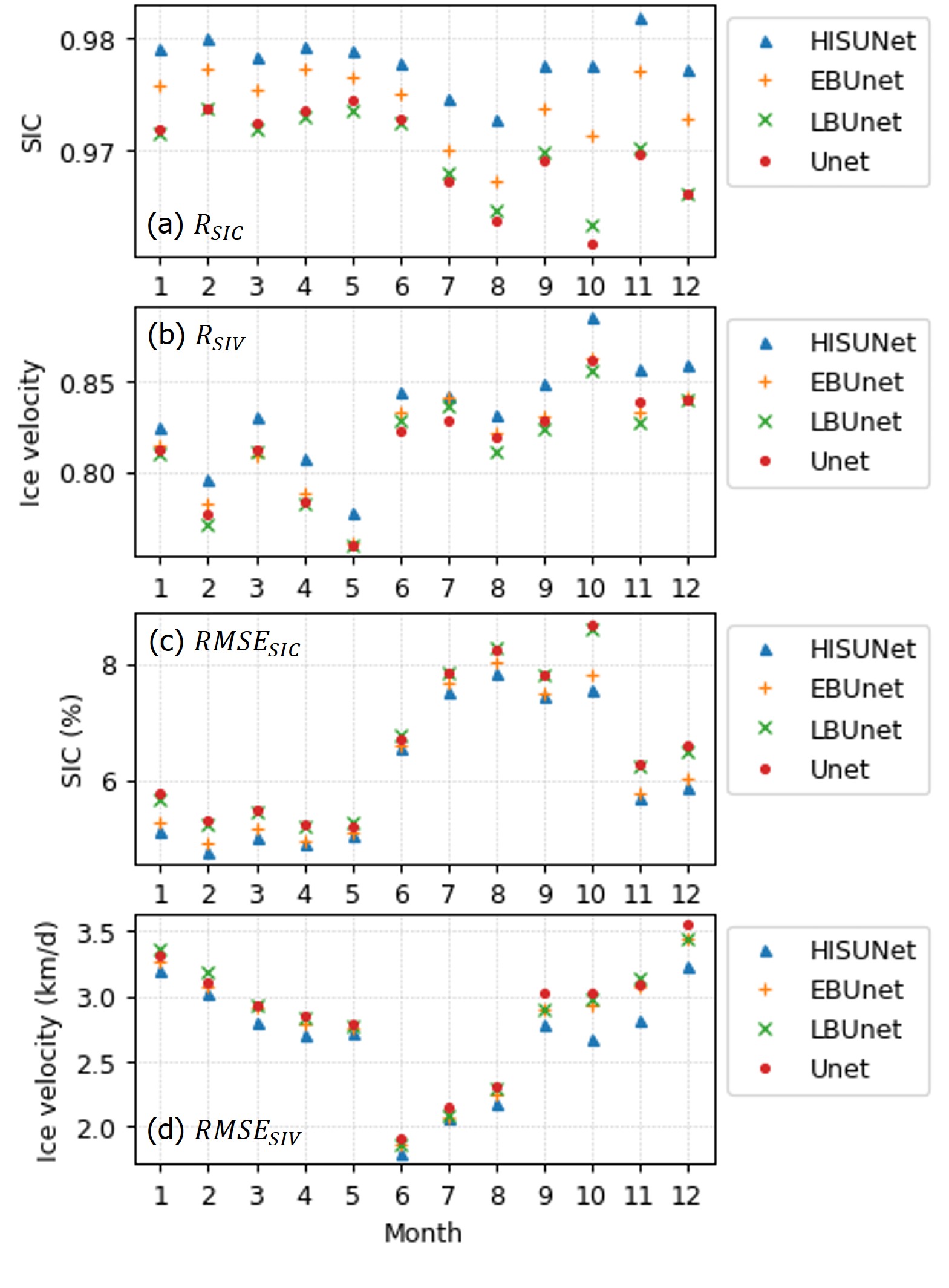}
    \caption{Comparison of monthly accuracy of the models in 2022: (a) R of SIC, (b) R of SIV, (c) RMSE of SIC, and (d) RMSE of SIV}
    \label{Error_by_month}
\end{figure}

\begin{figure}
    \centering
    \includegraphics[width=1.0\linewidth]{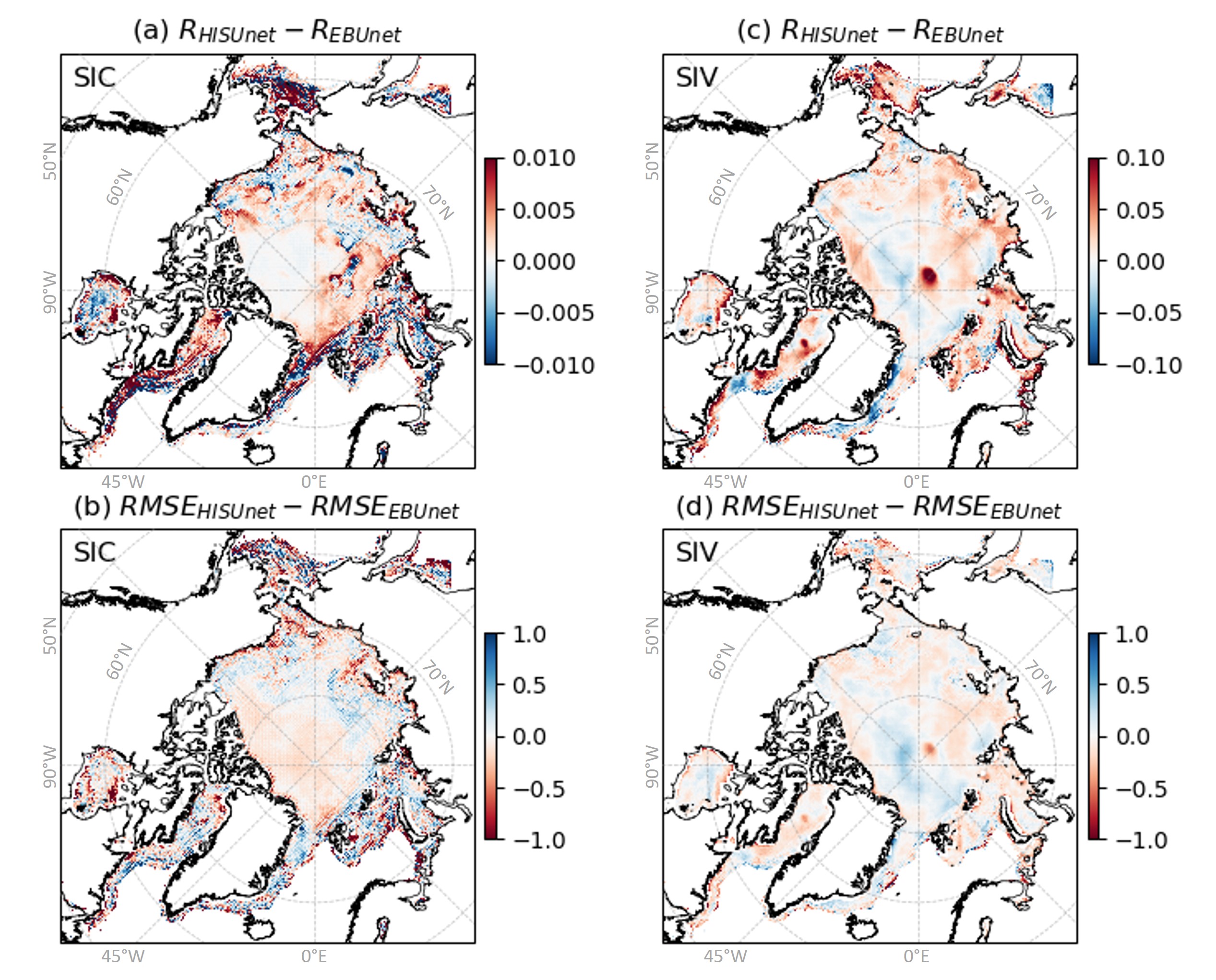}
    \caption{The difference of R and RMSE between HIS-Unet and EB-Unet: (a) R of SIC, (b) RMSE of SIC, (c) R of SIV, and (d) RMSE of SIV. All pixel values are the average of all days of 2022 when the SIC and SIV values of the pixels are valid.}
    \label{HISUnet-LBUnet}
\end{figure}

Fig. \ref{Error_by_month} depicts the monthly accuracy for each model. It is notable that the HIS-Unet has remarkably better performance during all months in both SIC and SIV prediction. In particular, the correlation coefficient derived by HIS-Unet outperforms the other machine learning models in both SIC and SIV prediction, which supports that the HIS-Uent produces the spatiotemporal variations of SIC and SIV more successfully (Fig. \ref{Error_by_month}a and \ref{Error_by_month}b). In terms of SIC prediction, the HIS-Unet shows a higher R by 0.002-0.006 and lower RMSE by 0.05-0.25 \%  compared to the EB-Unet. In particular, the largest improvement occurs in the summer months from July to October. While R varies significantly from July to October by other models ($<$ 0.97), the HIS-Unet keeps R $>$ 0.97 in these months, maintaining a similar level of R to winter months. October shows the most conspicuous improvement in SIC, reducing RMSE by 0.26 \% and increasing R by 0.006. In general, the Arctic sea ice extent decreases rapidly in July and reaches the minimum in September. Then, sea ice starts to grow from September, recording the fastest ice growth rate in October \cite{sea_ice_index}. As a result, the thermodynamic and dynamic sea ice condition changes more abruptly in the summer months. Integrating SIV information into SIC prediction via HIS-Unet might contribute to predicting SIC in this fast-melting and fast-growing season. Furthermore, as shown in Fig. \ref{HISUnet-LBUnet}a, the improvement of R is observed near marginal sea ice zones, rather than the central Arctic. This implies that SIV information can help predict spatiotemporal trends of SIC for where SIC changes dynamically in melting and freezing seasons. On the other hand, regarding the RMSE value of the SIC prediction, the HIS-Unet also contributes to reducing SIC errors slightly in the central Arctic (Fig. \ref{HISUnet-LBUnet}b). 

As for the SIV prediction, the R of HIS-Unet is higher than the EB-Unet by 0.01-0.02 and the RMSE is lower by 0.05-0.25 km/day on average. In the case of SIV, the largest improvement by HIS-Unet occurs in October-December: RMSE decreases more than 0.2 km/day, and R increases by 0.02. During these months, the Arctic sea ice extent starts to increase and sea ice becomes compacted. Such a compacted sea ice pack can either regulate or accelerate sea ice movement. Consequently, the integration of SIC information via HIS-Unet improves the SIV prediction when sea ice compactness increases in October-December. As shown in Fig. \ref{HISUnet-LBUnet}c and \ref{HISUnet-LBUnet}d, the improvement of SIV prediction occurs in marginal sea ice zones out of the central Arctic. Since the north of Greenland and the Canadian Archipelago are mainly covered by multi-year ice (MYI) that survives even in summer \cite{Meier2022}, the SIC in the central Arctic does not change much over a year. Although the integration of SIC information via the HIS-Unet does not efficiently improve the SIV predictability in such a static region of the central Arctic, it improves the SIV prediction broadly for the marginal ice zones where SIC changes seasonally.

\section{Discussion}\label{discussion}

Considering that the interaction between SIV and SIC can vary by sea ice conditions (e.g., marginal sea ice zones v. central Arctic), it is worth comparing the model accuracy for different regions. Thus, in the following subsection \ref{discussion1}, we discuss how the model accuracy changes for different regions of the Arctic Ocean. Furthermore, in subsection \ref{discussion2}, we examine the characteristics of information sharing between SIV and SIC layers, which occurs in WAMs. Finally, in subsection \ref{discussion3}, we discuss the climatological representativity of 2022 with respect to its impact on the training and testing of machine learning models. 

\subsection{Model accuracy for different Arctic subregions}\label{discussion1}

\begin{figure}
    \centering
    \includegraphics[width=1.0\linewidth]{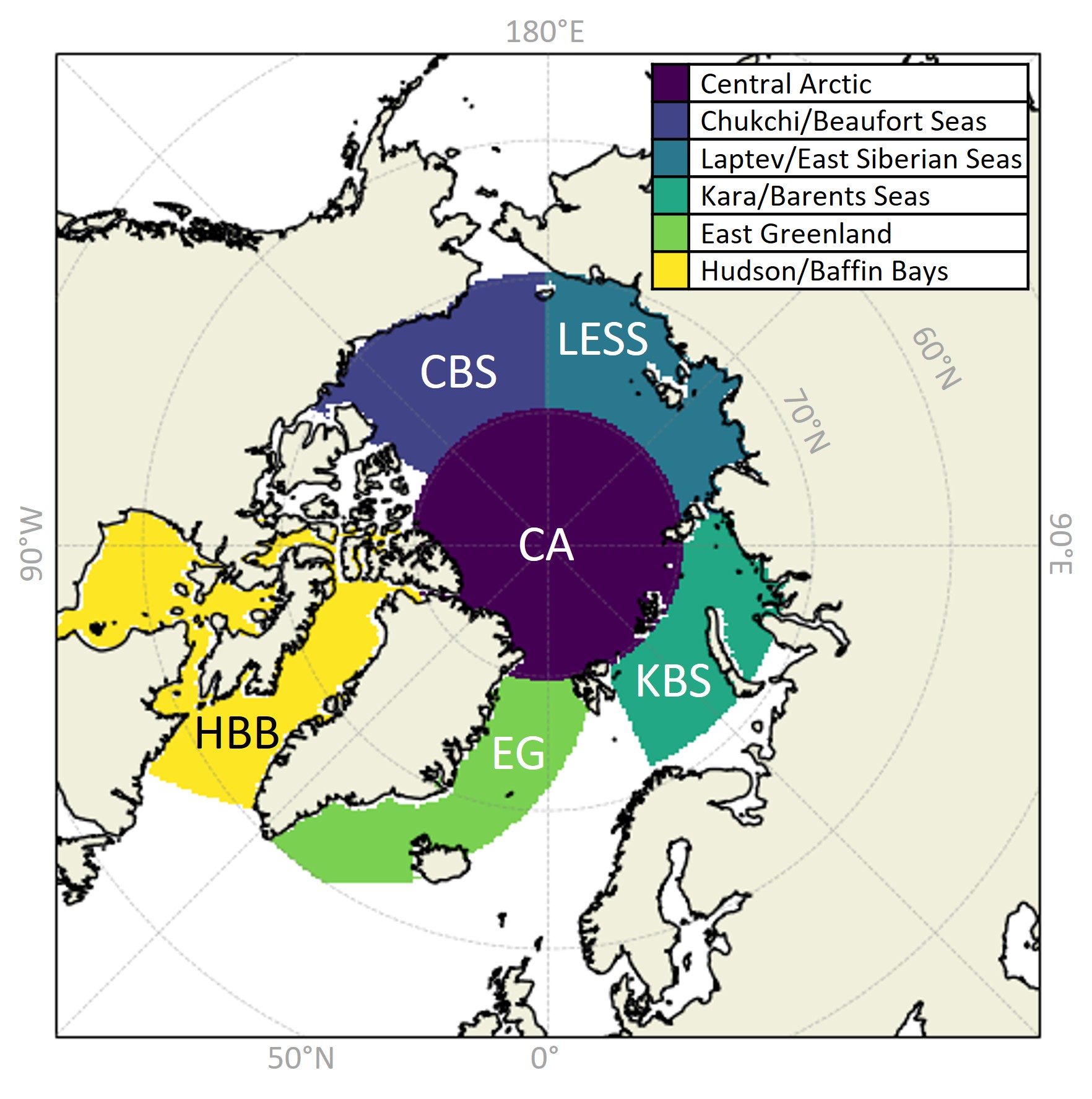}
    \caption{Six Arctic subregions defined in this study: Central Arctic (CA), Chukchi/Beaufort Seas (CBS), Laptev/East Siberian Seas (LESS), Kara/Barents Seas (KBS), East Greenland (EG), Hudson/Baffin Bays (HBB)}
    \label{Arctic_regions}
\end{figure}

\begin{figure}
    \centering
    \includegraphics[width=1.0\linewidth]{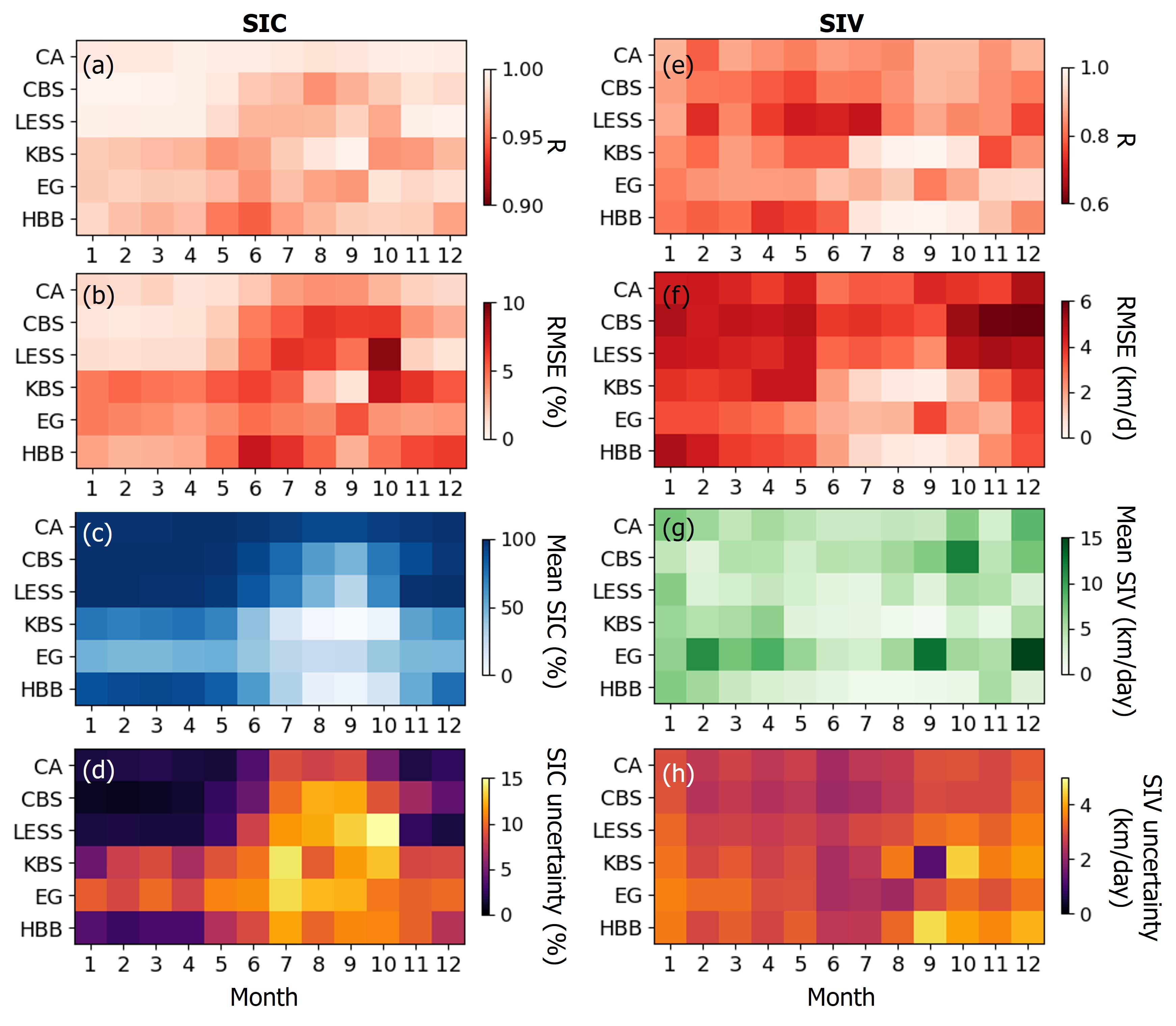}
    \caption{SIC prediction accuracy by subregions and months: (a) R of SIC and (b) RMSE of SIC. Bright white indicates better accuracy and dark red indicates worse accuracy. The monthly mean and uncertainty of SIC are displayed in (c) and (d), respectively. SIV prediction accuracy by subregions and months: (e) R of SIV and (f) RMSE of SIV. The monthly mean and uncertainty of SIV are displayed in (g) and (h), respectively.}
    \label{Error_by_region}
\end{figure}

We assess the accuracy of HIS-Unet for six different subregions in the Arctic Ocean (Fig. \ref{Arctic_regions}): Central Arctic (CA), Chukchi and Beaufort Seas (CBS), Laptev and East Siberian Seas (LESS), Kara and Barents Seas (KBS), East Greenland (EG), and Hudson and Baffin Bays (HBB). The division of these Arctic subregions has been conducted in many previous studies on the basis of their unique atmospheric and sea ice characteristics \cite{Parkinson1999, Overland2007, Cabral2022, Walsh2022, Chen2023}. Fig. \ref{Error_by_region} shows the RMSE and R of SIC and SIV prediction for these six subregions every month in 2022. 

Regarding the SIC prediction, CA shows the most accurate prediction, with the highest R and lowest RMSE for all months. This should be because the CA region is almost covered by high-concentration sea ice all months (Fig. \ref{Error_by_region}c). Since SIC remains stable at almost 100 \% for all seasons, the prediction of SIC in this region is relatively easier than in other regions. All other subregions also show R values greater than 0.95 and RMSE less than 10 \% for all seasons. The CA, CBS, and LESS regions show a seasonal trend in the SIC predictability: the RMSE increases in the summer months from June to October. The lower SIC prediction performance in the summer months might be because SIC changes significantly in these regions during the summer months, accompanied by fast melting and fast growing (Fig. \ref{Error_by_region}c). When these three regions are entirely covered by sea ice from January to April, the RMSE of SIC is less than 2 \% (Fig. \ref{Error_by_region}b). KBS shows a low error of SIC from August to September because of its almost ice-free condition during these months, but accuracy deteriorates in the fast ice-growing season in October-November and melting season in May-July. HBB also shows a similar accuracy variation with higher errors in melting seasons from May to July. EG shows relatively stable accuracy in SIC with R $>$ 0.97 and RMSE $<$ 5 \%, along with a consistent SIC all over the seasons. 

Next, in terms of the SIV prediction, it is interesting that the prediction performance is relatively stable in the CA region. While the other regions experience large fluctuations in R and RMSE values, the fluctuation of R and RMSE in CA is not as large as the other regions. The R-value remains 0.8-0.9 and RMSE 3.0-5.0 km/day for most of the months in CA. In CBS and LESS, the RMSE of SIV does not exceed 4.5 km/day before October, but RMSE increases up to 6 km/day in October-December. The KBS and HBB regions show abnormally high R and low RMSE values from July to October, but these values can be caused by very low SIC (Fig. \ref{Error_by_region}c) and slow sea ice drift (Fig. \ref{Error_by_region}g). Compared to the other subregions, EG shows extremely fast SIV (Fig. \ref{Error_by_region}c) as a gateway to a magnificent amount of sea ice exports through the Fram Strait \cite{Kwok2009}. Despite such fast SIV, this region shows relatively better accuracy than the other subregions, with $<$ 4 km/day of RMSE except for September.

Although our HIS-Unet model shows significant performance in both SIC and SIV prediction, here we note that the uncertainties in the target SIC and SIV observations should be taken into account. We calculate the monthly uncertainties in SIC and SIV observations for each subregion (Fig. \ref{Error_by_region}d and \ref{Error_by_region}h), which can be obtained from the SIC and SIV datasets \cite{Tschudi2020, NSIDC}. The uncertainties of SIC are relatively low in CA, CBS, and LESS compared to those in KBS, EG, and HBB, and the SIC uncertainties increase substantially in the summer months in all subregions. On the other hand, the regional and seasonal variations in SIV uncertainties are not as significant as those of SIC. However, KBS and HBB show slightly larger SIC uncertainties compared to the other regions from August to October, when sea ice melts and grows fast.

\subsection{Characterization of information sharing pattern in weighting attention modules}\label{discussion2}

\begin{figure*}[b]
    \centering
    \includegraphics[width=0.88\textwidth]{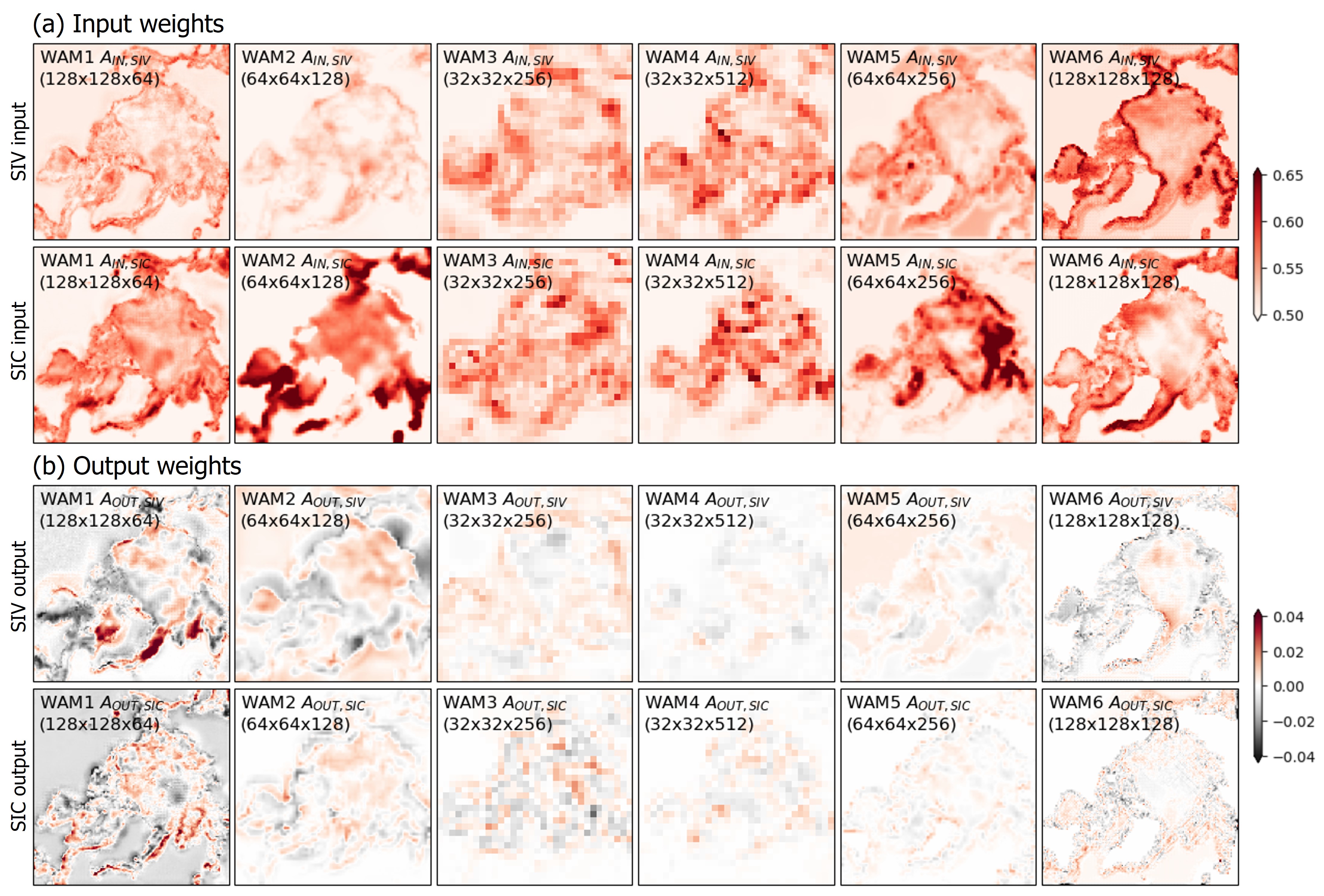}
    \caption{(a) WAM input weights ($A_{in,SIV}$ and $A_{in,SIC}$; refer to Fig. \ref{WAM_block} and Equation \ref{eq:4}) and (b) WAM output weights ($A_{out,SIV}$ and $A_{out,SIC}$; refer to Fig. \ref{WAM_block} and Equations \ref{eq:6_1} and \ref{eq:6_2}) for each level of HIS-Unet in Fig. \ref{Model_architectures}a.}
    \label{WAM_weights}
\end{figure*}

In the HIS-Unet architecture, the WAMs make it possible for the SIC and SIV prediction layers to share their information with each other. In the WAM, the information sharing is implemented by linear weight combination between SIC and SIV layers followed by channel and spatial attention modules. Moreover, the output feature from this WAM is passed through the multiplication of the output weights before being transferred to each branch. These input and output weights determine how much SIC and SIV information is to be mixed with each other and how significant this shared information is to be for each branch. Thus, we extract and examine the weight metrics of each WAM.

The input and output weights for WAM levels 1-6 are displayed in  \ref{WAM_weights}a and \ref{WAM_weights}b, respectively. Since each weight has the same dimension as the input layer ($H \times W \times C$), here we calculate the mean weights along all the channels ($H \times W$). In this section, we discuss the weighting values and spatial patterns exhibited only in the WAM1, WAM2, WAM5, and WAM6 because their $H \times W$ dimensions are closer to the original input and output dimension ($256 \times 256$). There is a possibility that other intermediate WAMs represent fine-tuning the weights and biases in the model rather than weighted information sharing.

First, when the SIV and SIC information are first blended through WAM1, SIC information is more weighted than SIV with higher input weights. In particular, interesting is that a higher $A_{in,SIC}$ is observed near the EG and HBB regions. These regions are known to have somehow different characteristics of sea ice dynamics from the other regions; HBB is relatively isolated from global oceanic currents in the central Arctic (e.g., Transpolar drift and Beaufort Gyre), and EG is a highly advective region (see the fastest ice velocity in Fig. \ref{Error_by_region}f) because of the fast sea ice near the Fram Strait. A higher $A_{in,SIC}$ in these regions implies that SIC information is useful to predict such distinctive SIV characteristics. When this shared information is transferred to each branch, the shared information in the HBB, EG, and KBS regions is weighted for the SIV branch (i.e., shown as higher $A_{out,SIV}$), and the shared information in the EG and KBS regions is weighted for the SIC branch (i.e., shown as higher $A_{out,SIC}$). We emphasize that the shared information is highly weighted in EG for both SIV and SIC predictions, which suggests that the information sharing between SIV and SIC is beneficial for predicting SIV and SIC in extremely dynamic regions like EG. In general, the magnitudes of output weights are higher in the SIV branch compared to the SIC branch.

At the next level of information sharing at WAM2, SIC information is also more weighted in sharing information, and the SIV branch receives more shared information in transferring the shared information. When the WAM2 receives information from SIV and SIC branches, SIC information is significantly weighted near marginal sea ice zones out of the central Arctic, particularly in KBS and HBB (i.e., higher $A_{in,SIC}$). This similar pattern is also observed in WAM5: more SIC contribution to shared information, particularly near KBS. Given that the variation of SIC in KBS is significantly influenced by the inflow and outflow of sea ice from the central Arctic \cite{Kwok2009}, the collaboration of SIC and SIV information can improve the model predictability in this region.

At the last information sharing state (WAM6), it is noted that the SIV information is weighted along the coastline in the information sharing (i.e., high $A_{in,SIV}$ values). The SIC information is also weighted more in the coastal regions (i.e., high $A_{in,SIC}$ values), especially near the EG region, than in the central Arctic. This indicates that the SIC information significantly contributes to predicting SIV in the EG region, in agreement with the results from WAM1. When the shared information transfers to each branch, the CBS and EG regions have higher $A_{out,SIV}$, while the magnitude of $A_{out,SIV}$ in the central Arctic is close to zero. The spatial variability of $A_{out,SIC}$ in WAM6 appears negligible.

In summary, we find the following noticeable characteristics in information sharing through WAMs of HIS-Unet: (1) information sharing between SIC and SIV is relatively active in sea ice marginal zones compared to the central Arctic; (2) when WAMs receive information from each branch and share them, SIC information is more weighted; (3) when WAMs pass the sharing information to each branch, the shared information is more weighted for the SIV branch than the SIC branch; (4) the information sharing between SIC and SIV is helpful to predict SIV in the HBB, EG, and KBS regions that have distinctive sea ice dynamics characteristics.

\subsection{Climatological representability in 2022}\label{discussion3}

\begin{figure}
    \centering
    \includegraphics[width=0.9\linewidth]{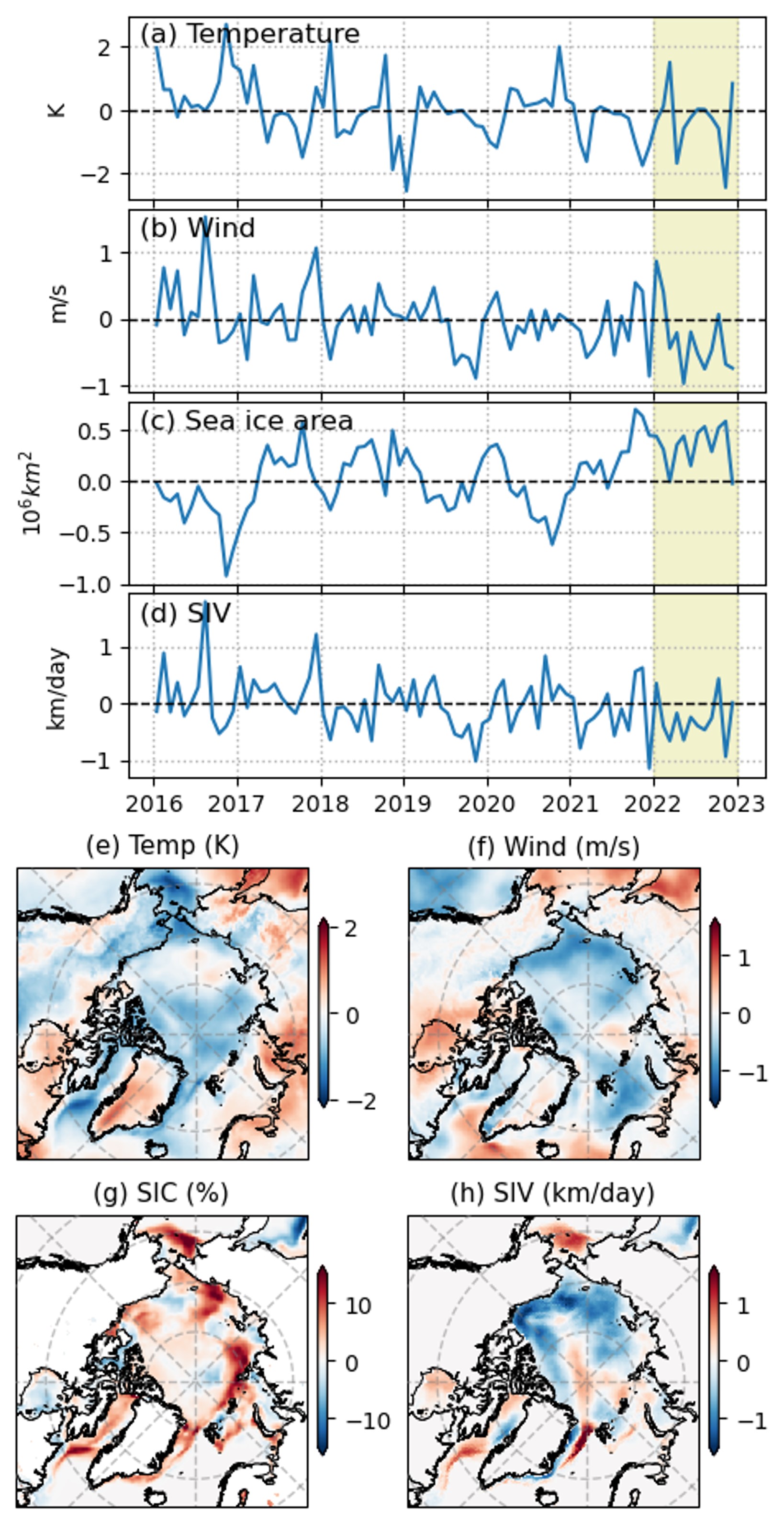}
    \caption{Monthly anomaly of (a) air temperature, (b) wind speed, (c) sea ice area, and (d) SIV to the 2016-2021 monthly average in the Arctic Ocean. Sea ice area is calculated as the area where SIC $>$ 15 \%. Anomaly maps of (e) air temperature, (f) wind speed, (g) SIC, and (h) SIV in 2022 to the 2016-2021 average.}
    \label{climatology}
\end{figure}

In this study, we use 2016-2021 data as training datasets and 2022 data as test datasets. If the climatological condition of 2022 is different from the other years, it can possibly make the model performance biased to the training datasets and reduce the representability of the model. Hence, in this section, we check if any climatological difference exists between training and testing years and how it affects the model performance. We compare the monthly anomalies of air temperature, wind speed, SIC, and SIV in 2022 to the average in 2016-2021. As shown in Fig. \ref{climatology}, 2022 showed lower temperature (Fig. \ref{climatology}a), lower wind speed (Fig. \ref{climatology}b), greater sea ice area (Fig. \ref{climatology}c), and slower SIV than 2016-2021. The lower anomalies of air temperature and wind speed are observed across the entire Arctic Ocean (Fig. \ref{climatology}e and \ref{climatology}f). SIC anomaly also occurs in most Arctic regions; in particular, LESS and KBS regions have extremely positive anomalies with $>$ 10 \% higher than the 2016-2021 average (Fig. \ref{climatology}g). Such SIC anomalies in LESS and KBS regions could bring relatively higher RMSE of SIC in these regions (Fig. \ref{Error_by_region}b). The slow SIV in CBS and LESS (Fig. \ref{climatology}h) appears somehow related to the negative anomaly of wind speed in these regions (Fig. \ref{climatology}f).

Herein, we emphasize that our HIS-Unet model shows significant performance in such abnormal climate and sea ice conditions in 2022 (as shown in Table \ref{table2} and Figs. \ref{Error_by_month} and \ref{Error_by_region}), even though the model is trained only with 2016-2021 data. Interesting is that the spatial distribution of SIC anomaly (Fig. \ref{climatology}g) is somehow overlapped with where HIS-Unet shows higher R than the EB-Unet (Fig. \ref{HISUnet-LBUnet}a). Introducing SIV information into SIC prediction could be beneficial to predicting the SIC anomaly condition in 2022. Regarding SIV, it is worth mentioning that the negative SIV anomaly regions in CBS and LESS show a slight improvement of SIV prediction by HIS-Unet (Fig. \ref{HISUnet-LBUnet}c and \ref{HISUnet-LBUnet}d).

\section{Conclusion}

In this study, we propose a multi-task fully convolutional network to predict Arctic sea ice concentration (SIC) and sea ice velocity (SIV) simultaneously. We accomplish the information sharing between SIC and SIV layers by adding six weighting attention modules (WAMs) between separate SIC and SIV U-net branches. The important SIC and SIV information is shared and highlighted in WAMs by following procedures: (1) Intermediate SIC and SIV features are mixed with multiplying linear input weights; (2) The shared information is highlighted through channel and spatial attention modules; (3) The highlighted shared information is transferred to SIC and SIV branches with multiplying linear output weights respectively. Since SIC and SIV affect each other in thermodynamic and dynamic ways, the information sharing and attention through WAMs can indirectly allow the network to learn their complicated interactions.

Our hierarchical information-sharing U-net (HIS-Unet) shows better performance than other network architectures, including convolutional network, U-net, late-branched U-net (LB-Unet), and early-branched U-net (EB-Unet). In particular, our HIS-Unet outperforms these models in the melting or early freezing seasons and shows a stable performance in the central Arctic, where the SIC is relatively consistent throughout the year. For the other regions, although the model performance varies by the sea ice conditions, the fidelity of HIS-Unet is close to or better than the previous SIC and SIV prediction studies. When the input and output weights of WAMs are examined, we find that the information-sharing scheme of HIS-Unet plays an important role in predicting SIC and SIV in the marginal sea ice zones where sea ice conditions change dynamically, such as East Greenland, Hudson/Baffin Bays, and Krara/Barents Seas. In these regions, WAMs explicitly contribute to SIC and SIV predictions by (i) assigning higher weights to SIC information in these regions when mixing SIC and SIV information and (ii) assigning higher weights to shared information in these regions when transferring shared information to the SIC and SIV branches. Furthermore, the information sharing between SIC and SIV also helps predict the anomalous sea ice conditions in 2022, even though this model is trained without this year's data.

\section*{Acknowledgments}
This work is supported by NSF BIGDATA (IIS-1838230, 2308649) and NSF Leadership Class Computing (OAC-2139536) awards. 

\bibliographystyle{ieeetr}
\bibliography{references}

\vspace{-0.7cm}

\begin{IEEEbiography}[{\includegraphics[width=1in,height=1.25in,clip,keepaspectratio]{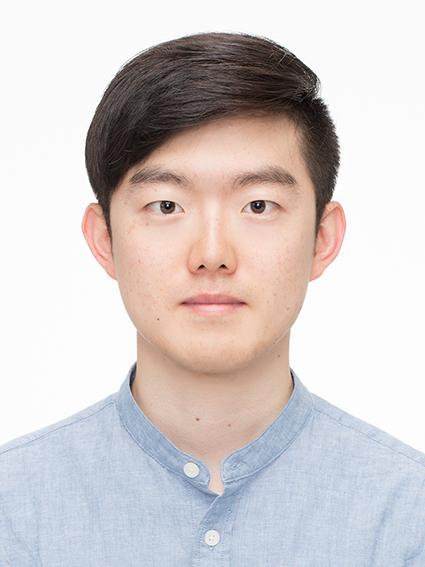}}]{Younghyun Koo} received the B.S. degree in Energy resources engineering from Seoul National University, Republic of Korea, in 2017, the M.S degree in Energy systems engineering from Seoul National University, Republic of Korea, in 2019, and the Ph.D degree in Environmental Science and Engineering from the University of Texas at San Antonio, TX, USA, in 2023. He is currently a postdoctoral research associate at the Computer Vision and Remote Sensing Laboratory (Bina lab) at Lehigh University. His research interest is the application of machine learning and remote sensing techniques for environmental monitoring, particularly in polar regions.
\end{IEEEbiography}

\vspace{-0.7cm}

\begin{IEEEbiography}[{\includegraphics[width=1in,height=1.25in,clip,keepaspectratio]{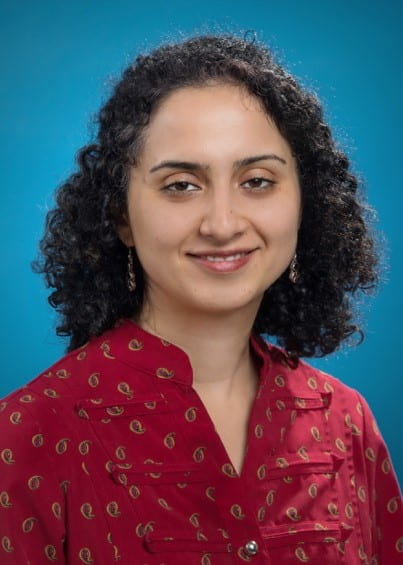}}]
{Maryam Rahnemoonfar} received her Ph.D.
degree in Computer Science from the University of
Salford, Manchester, U.K. She is currently an Associate Professor and Director of the Computer Vision and Remote Sensing Laboratory (Bina lab) at Lehigh University. Her research interests include Deep Learning, Computer Vision, Data Science, AI for Social Good, Remote Sensing, and Document Image Analysis. Her research specifically focuses on developing novel machine learning and computer vision algorithms for heterogeneous sensors such as Radar, Sonar, Multi-spectral, and Optical. Her research has been funded by several awards including the NSF HDR Institute award-iHARP, NSF BIGDATA award, Amazon Academic Research Award, Amazon Machine Learning award, Microsoft, and IBM. 
\end{IEEEbiography}


\vfill
\end{document}